# Wavelength-multiplexed Delayed Inputs for Memory Enhancement of Microring-based Reservoir Computing

**Bernard J. Giron Castro,**[1*] **Christophe Peucheret,**[2] **and Francesco Da Ros**[1]
[1]*DTU Electro, Technical University of Denmark, 343 Ørsteds Plads, DK-2800 Kongens Lyngby, Denmark*
[2]*Univ Rennes, CNRS, UMR6082 - FOTON, 22305 Lannion, France*
[*]*bjgca@dtu.dk*

**Abstract:** We numerically demonstrate a silicon add-drop microring-based reservoir computing scheme that combines parallel delayed inputs and wavelength division multiplexing. The scheme solves memory-demanding tasks like time-series prediction with good performance without requiring external optical feedback. © 2023 The Author(s)

## 1. Introduction

Photonic technologies are promising for boosting the performance of current computing architectures [1, 2]. Among their benefits is the potential for parallel processing using mature technologies such as wavelength division multiplexing (WDM) which has been applied to photonic neural networks [2]. In our previous study [3], we showed the potential of WDM to improve the computing capabilities of a reservoir computing (RC) scheme. RC is a relatively recent computing paradigm that uses random fixed weights and complex nonlinear dynamics to map the input data to a higher dimensional space. This process allows simplifying its training process, which is only required in its output layer (usually through linear or ridge regression). The performance of RC is highly related to the nonlinear dynamics of the nodes and the capability to buffer past inputs (memory) [1]. Nevertheless, photonic RC schemes may reduce their scalability if multiple photonic blocks are implemented as nonlinear nodes. An approach to address this issue is time-delay RC (TDRC), in which the (virtual) nodes are spatially and thus temporally distributed. Each virtual node is processed, one at a time, through a single physical nonlinear node, minimizing the hardware requirements [1].

Previous photonic TDRC implementations [3-6] have demonstrated that a silicon MRR can perform as the nonlinear node of an TDRC scheme when exploiting free-carrier dispersion and thermo-optic nonlinear effects. These nonlinear effects provide very short memory by themselves [6]. Therefore, a mechanism to enhance the memory capacity while not reducing the speed and scalability of the photonic TDRC is required [1]. To provide the necessary increase in memory and spatial distribution, the addition of delayed feedback by means of an external waveguide has been studied [4, 5]. However, the addition of the feedback delay is rather challenging for photonic integration. Additionally, due to the propagation of the optical signal through the feedback, the phase control and the compensation for losses are potential concerns in practical implementations.

In this work, we use four optical frequency modes modulated by copies of the input sequence, each with a different but systematic amount of added delay. We demonstrate that it is possible to remove the physical delayed feedback in an MRR-based TDRC scheme while still achieving similar performance levels in a memory-demanding task such as NARMA-10. This offers the potential to improve the scalability of MRR-based TDRC schemes.

## 2. Setup and methodology

The proposed scheme is shown in Fig 1. The nonlinear dynamics of the MRR considering $M$ optical pumps follow the temporal coupled-mode theory model described in [3], which is solved using a 4$^{th}$-order Runge-Kutta method with a step $\eta = 2.0$ ps. The $M$ copies of the same input sequence $u(n)$ linearly modulate each optical pump, where each $i^{th}$ copy has an added delay $\tau_i = i \cdot S_w/\eta$ over the discretized sequence (in terms of the RK solver). $S_w = 1.0$ ns is the length of the symbol. Every sequence is multiplied by its respective (different seed) masking signal $m_i(n)$ with size $N = 50$ based on the delay input method studied in [7]. Each masking sequence is taken from a uniform distribution over the interval [0, 0.5] and determines the number of virtual nodes. An optimized task-independent bias $\beta$ is added to their product, as done in [3]. The resulting signal modulates the intensity of its respective optical carrier before being injected into the MRR using a WDM combiner. Notice that no external feedback is added to the MRR in this proposed scheme. At the drop port, the $M$ optical signals are wavelength-demultiplexed and detected by individual photodiodes. Subsequently, the sum of the detected signals is obtained. The computing process is described in Eq. (1), where $\mathcal{F}$ represents the optoelectronic nonlinear transformation of each input sequence when processed through the MRR and $X(n)$ is the calculated response of the RC. The optical parameters of the setup are defined in [3, 4]. We consider the NARMA-10 task but the potential of MRR-based RC has been demonstrated for other time-series prediction tasks [5]. 3200 data points are used for training and 800 for testing. $X(n)$ is then used to train the weights via ridge regression with a regularization parameter $\Lambda = 1\times10^{-9}$. The performance of the system is evaluated using the normalized mean square error (NMSE) between the predicted and target sequences.

$$X(n) = \sum_{i=0}^{M} \mathcal{F}[u_i(n-\tau_i)m_i(n) + \beta] \quad (1)$$

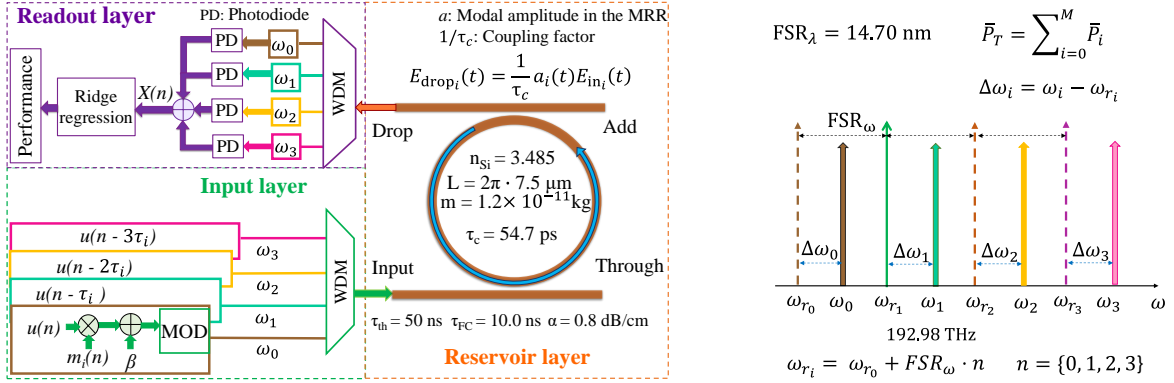

Fig. 1. Proposed RC scheme and frequency allocation of the four optical pumps.

## 3. Results

The system is simulated for a total average input power $\bar{P}_T$ [-15, +25] dBm equally distributed between the four optical pumps ($M = 4$ as shown in Fig. 1), each with an equal frequency detuning $\Delta\omega/2\pi$ within a ±100 GHz range from its respective resonance. The obtained results are shown in Fig. 2a. For comparison, we also present the results obtained for a single wavelength ($M = 1$) when no external feedback is applied between the through and add ports of the MRR (Fig. 2(b)), as well as when such feedback is added (Fig. 2c). In every case, the same task is solved using the same optical and computing parameters. Minima and maxima obtained NMSEs are displayed as extreme values of the colorbar legends. The proposed approach exhibits performance comparable to the single-λ system with feedback [4] (Fig. 2c) and significantly better than the single-λ without feedback. Furthermore, the RC with WDM delayed inputs mantains relatively good performance over a broader range of $\Delta\omega$, even where the MRR appears to work in a linear regime [4]. The power required for the best performance, though, is higher for the system under investigation, but it is important to note that very low attenuation was considered when adding an external waveguide in [4].

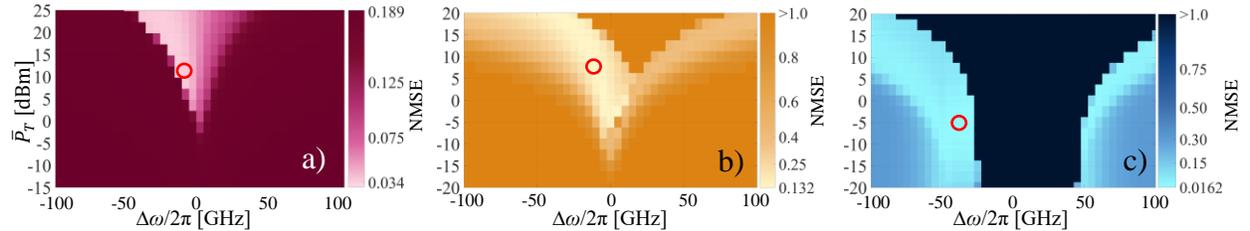

Fig. 2. NMSE as a function of $\bar{P}_T$ and $\Delta\omega/2\pi$ for: a) RC with WDM delayed inputs. b) Single-λ RC scheme without delayed feedback. c) Single-λ RC scheme with delayed feedback. Red circles mark the best performance.

## 4. Conclusion

The potential of parallelizing delayed inputs to enhance the memory of RC schemes is shown. The proposed silicon MRR-based RC scheme improves its scalability as it does not require an external waveguide or optical fiber. This avoids potential issues related to the feedback while still achieving an NMSE <0.05. The system could be further improved by optimizing the input power and frequency detuning of each optical pump or by increasing $M$.

**Acknowledgment** Villum Fonden (VIL29334) and Vetenskapsrådet (2022-04798).